\documentclass[[12pt,onecolumn,journal]{IEEEtran}
\ifCLASSINFOpdf
\usepackage[pdftex]{graphicx}
\else
\fi
\usepackage{silence}
\usepackage{amsmath,amssymb,amsfonts}
\usepackage{mathtools}
\usepackage{subcaption}
\usepackage{float}
\usepackage{comment}
\usepackage{placeins}
\usepackage{cite}
\usepackage{xcolor}
\usepackage{adjustbox}
\usepackage{multirow}
\usepackage{hyperref}
\usepackage{array}
\usepackage{graphicx}
\usepackage{textcomp}
\usepackage{mathtools}
\usepackage{mathrsfs}
\usepackage{breqn}
\usepackage{array}
\usepackage{mathrsfs}

\DeclareMathOperator*{\argmax}{arg\,max}

\usepackage{braket}
\newcolumntype{P}[1]{>{\raggedright\arraybackslash}p{#1}}
\ifCLASSINFOpdf
\else
\fi
\begin{document}
\title{Implementation of a Three-class Classification LS-SVM Model for the Hybrid Antenna Array with Bowtie Elements in the Adaptive Beamforming Application}
\author{Somayeh~Komeylian, and~Christopher~Paolini,~\IEEEmembership{IEEE,~Member}}


\markboth{September~2022}%
{Shell \MakeLowercase{\textit{et al.}}: Bare Demo of IEEEtran.cls for IEEE Journals}
\maketitle
\begin{abstract}
To address three significant challenges of massive wireless communications including propagation loss, long-distance transmission, and channel fading, we aim at establishing the hybrid antenna array with bowtie elements in a compact size for beamforming applications.
In this work we rigorously demonstrate that bowtie elements allow for a significant improvement in the beamforming performance of the hybrid antenna array compared to not only other available antenna arrays, but also its geometrical counterpart with dipole elements.
We have achieved a greater than 15 dB increase in SINR values, a greater than 20\% improvement in the antenna efficiency, a significant enhancement in the DoA estimation, and 20 increments in the directivity for the hybrid antenna array with bowtie elements, compared to its geometrical counterpart, by performing a three-class classification LS-SVM (Least-Squares Support Vector Machine) optimization method.
The proposed hybrid antenna array has shown a 3D uniform directivity, which is accompanied by its superior performance in the 3D uniform beam-scanning capability. 
The directivities remain almost constant at 40.83 dBi with the variation of angle $\theta$, and 41.21 dBi with the variation of angle $\phi$.
The unrivaled functionality and performance of the hybrid antenna array with bowtie elements makes it a potential candidate for beamforming applications in massive wireless communications. 
\end{abstract}

\begin{IEEEkeywords}
Antenna array, SINR, Antenna arrays, Antenna efficiency, Support vector machine, Beamforming techniques.  
\end{IEEEkeywords}
\IEEEpeerreviewmaketitle
\section{Introduction}
\IEEEPARstart{R}{ecent} advances and innovative solutions have been extensively developed for deploying smart array antennas with a highly-directive radiation pattern in order to overcome the long-distance challenges and minimize interference signals, as well as enhance the spherical coverage and capacity in massive wireless communication channels. 
In this sense, in addition to performing optimization methods, the geometrical configuration, inter-element spacing, excitation phase, and amplitude of the individual array elements of each antenna array have been designed and synthesized for controlling its radiation patterns. 

A linear or one-dimensional antenna array is employed for applications in which a directive radiation pattern and broad coverage in a plane orthogonal to the antenna array is required.
In other words, users can be observed by the angle-of-arrivals (AoAs)~\cite{Liu}, in the azimuth domain (A-AoAs) of linear antenna arrays. 
As discussed in detail in~\cite{Yasar}, the uniform linear array (ULA) is employed for finding sources in front of the antenna array, for which it is assumed there is no source behind the antenna array. 

One way to overcome the aforementioned limitations consists in providing an extra degree of freedom in the elevation angle domain by utilizing the uniform rectangular array (URA). 
Hence, users can be detected not only by angle-of-arrivals in the azimuth domain, but also by angle-of-arrivals in the elevation domain (E-AoAs) of the URA.
However, the URA has demonstrated weak performance with a low resolution for the E-AoAs in the discrete angular domain, as discussed in~\cite{Mahmoud}. Another disadvantage of the URA consists of having an additional major lobe in the opposite side of its mainlobe with the same intensity.

The arrangement of the circular antenna array does not have any edge elements, thus the circular array represents more robustness to the mutual coupling effect compared to the ULAs and URAs ~\cite{Balanis-2012-antenna}. 
In this sense, the circular antenna array can mitigate the mutual coupling effect by breaking down the antenna array excitation into a series of symmetrical spatial components. 
Moreover, a directive radiation pattern synthesized by the circular antenna array can electronically rotate in its own antenna array plane without any significant change in its beam shape
~\cite{Balanis-2012-antenna}. 
Although a uniformly excited and equally-spaced arrangement in the circular antenna array provides high directivity, the circular array configuration suffers from high side lobe level (SLL). 
Grating lobes and sidelobes are two underlying causes of the vulnerability of the circular antenna array to noise and interference signals. 
Furthermore, the radiation pattern of the circular array does not include any no null in the azimuth plane, which is required for nullsteering in beamforming applications.

Two geometrical configurations of the uniform hexagonal array (UHA)~\cite{Mahmoud} and concentric circular antenna array (CCAA)~\cite{Ram} can drastically mitigate the SLL effect in the circular antenna array.
However, the high grating lobe, in which the sidelobe amplitude approaches the mainlobe level, still appears in the radiation pattern of the uniform concentric circular array~\cite{Noordin}. 
The hexagonal arrangement has demonstrated better steerability and gain with a lower SLL compared to the uniform and planar circular arrays~\cite{Mahmoud}. 

The 2D antenna array shows better performance in the azimuth direction, however its 3D geometrical counterpart has better performance in the elevation direction.
Although planar antenna arrays can significantly overcome propagation loss and thereby increase the high-frequency bandwidth of wireless communications by steering the radiation energy towards desired directions, they may not capable of supporting long-distance transmission and providing strong robustness to variations of wireless communication channels.
The performance of multi-layer elements or 3D array elements in terms of scanning angle, and thereby spatial resolution, can be improved by an appropriate choice of antenna~\cite{Harabi,Li,Hawkes}. 
Among the 3D antenna array geometries, the cylindrical antenna array, whose radiation pattern has several nulls in the azimuth plane, can significantly suppress interference signals and noise. 
Although the concentric circular sub-arrays in the cylindrical antenna array allow mitigating the SLL, the mutual coupling effect is still a severe challenge in the cylindrical antenna array~\cite{Hussain}.
Although the spherical antenna array has the capability of organizing the non-uniform elevation spacing of antenna elements on its 3D geometrical configuration, which is accompanied by a drastic sidelobe reduction and a super directive characteristic~\cite{Adeel,Elizarraras}, it has a complex array geometry.

While the planar antenna array becomes robust to variations of a wireless channel by performing optimization methods or beamforming techniques, its performance can still degrade in a massive wireless channel. 
In other words, the antenna array performance is less vulnerable to variations of a wireless channel when deeper nulls occur at angles of interference signals. 
Compared to available planar antenna arrays in~\cite{Komeylian2020}, the radiation pattern of the 3D conical antenna array consists of deeper nulls~\cite{Ghavami}, which is accompanied by its better performance in terms of suppressing interference signals.  



To overcome the limitations and technical imperfections of the available antenna arrays~\cite{Liu,Balanis-2012-antenna,Ram,Noordin,Mahmoud,Harabi,Li,Hawkes,Hussain,Adeel,Elizarraras,Komeylian2020,Ghavami}, this work's major contribution is implementing a three-class classification LS-SVM for the proposed hybrid antenna array to perform the adaptive beamforming technique and DoA estimation at the frequency of operation of 10 GHz.

The adaptive 3D hybrid antenna array with bowtie elements has to be capable in efficiently directing the maximum antenna gain towards desired signals and nulls towards interference signals. 
By adaptively updating weights of the array elements, the hybrid antenna array is capable of tracking desired signals and suppressing interference signals so that the maximum SINRs are achieved.   
We have demonstrated that a major advantage of the 3D antenna arrays consists in having greater values of SINRs. 
Higher values of SINRs guarantee the superiority of the antenna array performance for massive wireless communication channels with long-distance transmission. 

Noordin et al. in~\cite{Noordin} have performed the particle swarm optimization method for different planar circular arrays to increase their capabilities to adapt to the variations of massive wireless channels. 
They verified that a maximum value of SIR of 3.04 dB is associated with the planar uniform circular array, which cannot meet the requirements of long-distance transmission of a wireless communication channel. 

Ioannides et al. in~\cite{Ioannides} have implemented two different optimization methods of MUSIC and ESPRIT for enhancing the performance of the proposed circular antenna array for beamforming applications. They confirmed that the circular antenna array has a maximum SNR of 10 dB. 


In the 3D proposed hybrid antenna array, elements were placed at different geometrical locations, therefore, a different set of incident directions are observed by their terminals. 
This results in high variations in the correlation patterns of array elements across their terminals, yielding an unequal correlation matrix. 
Furthermore, in practice, the spatial correlation of the antenna array is significantly affected by the type of wireless channel: line-of-sight (LoS), and non-line-of-sight (NLoS). 
In the worst scenario, in which the wireless channel includes NLoS propagation and the spatial correlation matrix is unequal, the hybrid antenna array with bowtie elements has shown an SINR value of 31.565 dB.

Another advantage of antenna arrays, which allows them to be the potential candidate for beamforming applications in a massive wireless channel, consists of having a high directivity. 
Khalaj in~\cite{Khalaj-Amirhosseini} has synthesized different linear and planar antenna arrays to maximize the directivity parameter. 
He confirmed that the maximum achievable directivity for the planar antenna arrays is 25 dB. 


Yaacoub et al. in~\cite{Yaacoub} have demonstrated that the directivity of the cylindrical antenna array is significantly affected by the type of excitations. 
A maximum directivity of 14.87 dB is associated with the cylindrical antenna array excited uniformly~\cite{Yaacoub}.  


The deep neural network has been performed for enhancing the performance of the different antenna arrays under the same conditions in~\cite{9495844}.
The proposed hybrid antenna array with dipole elements~\cite{Komeylian2021} has demonstrated a superior directivity compared to the other available antenna arrays.  


The hybrid antenna array has demonstrated a directivity of more than 100 due to the effectiveness and strength of the proposed three-class classification LS-SVM algorithm. 

Consequently, in addition to the introduction section, this work is organized into the four following sections: section II is allocated for representing the proposed geometrical configuration of the hybrid antenna array proposed in this work. 
In section III, the involved techniques and approaches have been implemented for the hybrid antenna array for the DoA estimation and beamforming technique. 
Section VI consists of evaluating the proposed hybrid antenna array in terms of SINR, antenna efficiency, and uniform 3D-beam scanning capability in beamforming applications in wireless communication channels. 
Section V provides the conclusions of this work. 

\section{Hybrid Antenna Array with Bowtie Elements}

We aim to demonstrate the strong reinforcement characteristics and superior performance of the hybrid antenna array with bowtie elements over its geometrical counterpart with dipole elements and other available antenna arrays for the beamforming application in massive wireless communications. 

\begin{figure}[htbp]
(a)
\centering 
\includegraphics[width=0.52\textwidth,height=0.32\textwidth]{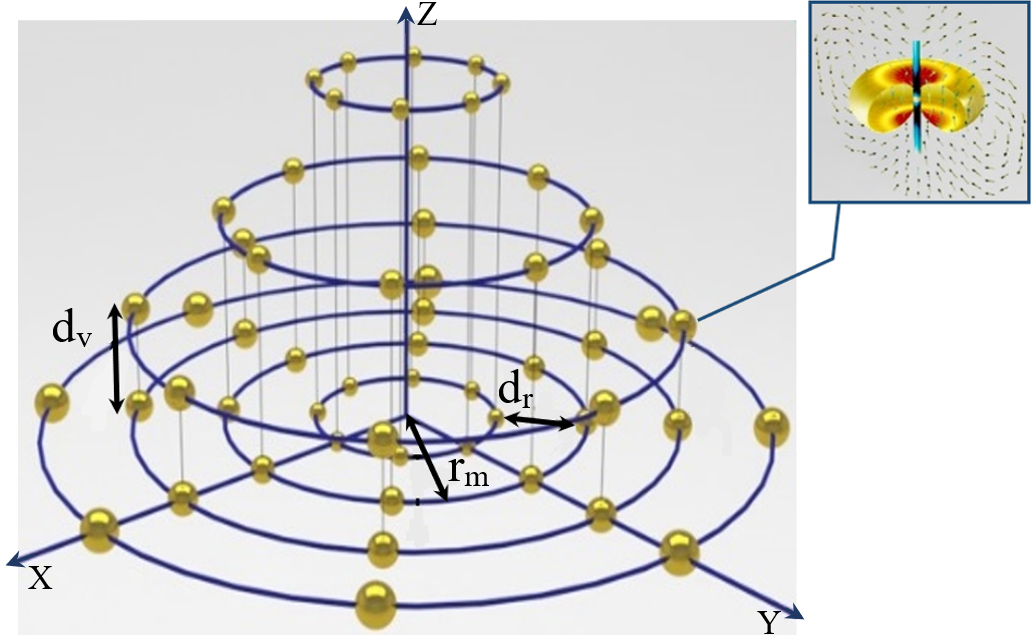}\\(b)
\includegraphics[width=0.52\textwidth,height=0.32\textwidth]{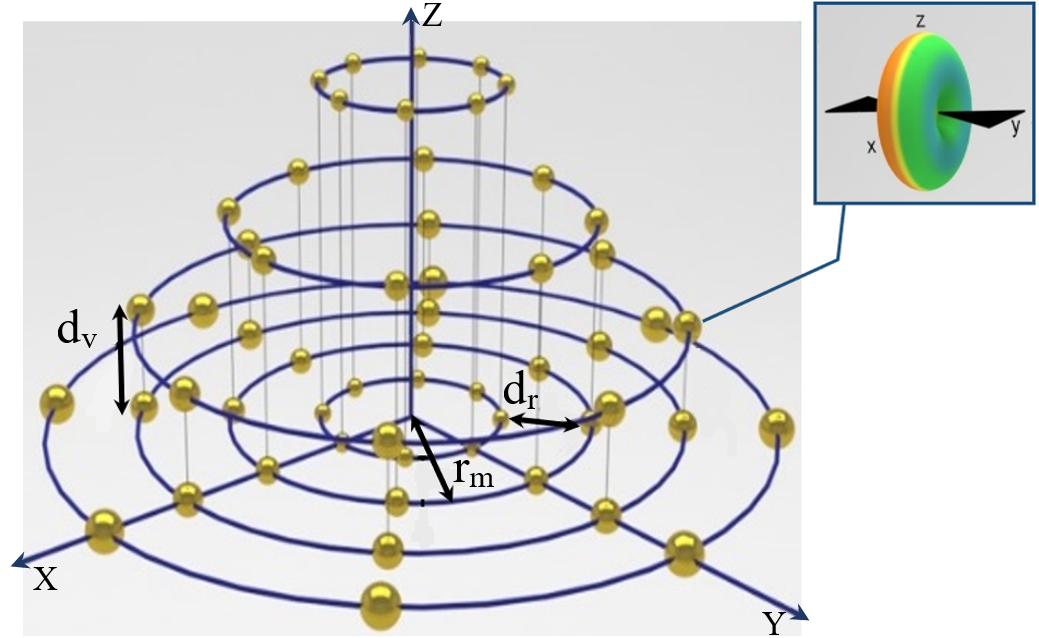}
\caption{Demonstration of the hybrid antenna array at the frequency of the operation of 10 GHz with two different array elements: (a) The 3D configuration of the hybrid antenna array with the dipole elements. The assigned design parameters of the hybrid antenna array are listed in detail in the Table 1, (b) The 3D configuration of the hybrid antenna array with the bowtie elements. The assigned design parameters of the hybrid antenna array are listed in detail in the Table 1. In this work, each hybrid antenna array consists of three cylindrical antenna arrays~\cite{Hussain}, and one circular antenna array~\cite{Noordin}.}
\label{fig:configuration}
\end{figure}

The proposed hybrid antenna array is composed of cylindrical sub-arrays. 
Each cylindrical sub-array includes two concentric circular sub-arrays positioned upon each other, as demonstrated in Fig.~\ref{fig:configuration}. 
The proposed 3D configuration allows for having advantages of both circular and cylindrical antenna arrays, simultaneously, in addition to reducing the size and the number of elements, which is comparable to that obtainable with an ideal case of a spherical array~\cite{Komeylian20212,Komeylian2021}.


\begin{figure}[htbp]
\centering
\includegraphics[width=0.25\textwidth,height=0.34\textwidth]{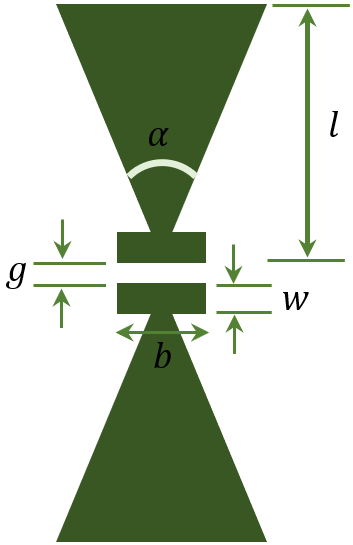}
\caption{Top view of the bowtie antenna; $l$: arm length, $\alpha$: flare angle, $g$: feed gap, $w$: bar width, $b$: bar length. The design parameters of each bowtie element in Fig.1(b) are assumed to be $l$=6 mm, $\alpha$=60°, $g$=0.02 mm, $w$=0.02 mm, and its thickness is equal to 0.01 mm.}
\label{fig:top-view}
\end{figure}

\begin{table}[ht]
\centering
    \begin{tabular}{c|c|c}
Parameters &  Definition & Value \\ \hline
        $N_h$ & Number of elements of any circular loop &	$N_h=20$ \\
        $Q_h$ & Number of elements of any cylinder & $Q_h=40$ \\
        $M_h$ & Total number of cylinders in the proposed array & $M_h=3$ \\
        $P_h$ & Number of circular loops in the cylinder & $P_h=2$ \\
        $d_v$ & Vertical spacing between two consecutive circular loops & $d_v=0.5\lambda$ \\
        $d_r$ & Horizontal spacing between two consecutive circular loops &	$d_r=0.5\lambda$ \\
        $\phi,\theta$ & Maximum scanning angles & $\phi=45^{\circ}, \theta=45^{\circ}$
    \end{tabular}
    \caption{The design parameters for the proposed hybrid antenna array.}
    \label{tab:parameters}
\end{table}
This work has a further contribution to verifying a noticeable difference between the performance of the hybrid antenna array with dipole elements compared to its geometrical counterpart with bowtie elements in the beamforming application for massive wireless communications. 
Every bowtie antenna is built of the two very small and conductive thin triangular segments, which are fed at the bow apex, as demonstrated in Fig.~\ref{fig:top-view}.

The geometric shape of the bowtie antenna is modified to enlarge its flare angle. 
The bowtie antenna with the longer arm length or the higher flare angle allows current flow through the longer path to the gap and thereby the effective size of the antenna is increased.
The hybrid antenna array with bowtie elements outperforms its geometrical counterpart with dipole elements and other available antenna arrays due to the four distinct features of its bowtie elements.  

\begin{itemize}
    \item The tapered shape of the bowtie antenna does not allow its VSWR performance to degrade as the frequency increases. Hence, the bowtie antenna has a better impedance manner across its bandwidth compared to its geometrical counterpart from the dipole antenna.

    \item The unidirectional bandwidth of the dipole antenna has the potential application for ultra-high resolution radar and modern wireless communications.  

    \item The geometrical shape of the bowtie antenna is adjustable such that the antenna area is enlarged by increasing its flare angle while its radius remains unchanged. 

    \item The skin effect in antennas is significantly affected by the frequency of the operation. However, as frequency increases from 300 MHz to 3 GHz, the skin effect does not drastically degrade the performance of the bowtie antenna. Therefore, we can achieve a better response by increasing the thickness of the bowtie antenna.
\end{itemize}

Consider the spatial distribution of the hybrid array elements in which the $nm$th element consists of $I_{mn}$ weightings and $\beta_{m}$phase shift. 
Therefore, the array factor can be expanded in terms of a superposition of the weighting vectors of all the individual elements as follows~\cite{Saqib2015AHA},
\begin{multline}
AF(\theta,\phi)=\underbrace{\sum_{n=1}^N I_n e^{j k r_0 \sin \theta \cos(\phi-\phi_n) +\alpha_n}}_{\text{for the circular antenna array}} 
+ \underbrace{\sum_{m=1}^M \sum_{p=1}^2
\sum_{n=1}^N I_{mn} e^{j(p-1)(kd_m \cos \theta + \beta_m)}  e^{jkr_m \sin \theta \cos(\phi_{mn}-\phi)}}_{\text{for the three cylindrical antenna arrays}}
\label{eq:af}
\end{multline}
\noindent where $I_{mn}$ refers to the current excitation of the individual array elements. 
For the uniform excitation, weights of the array elements are supposed to be a constant value of $I_{mn}=I_0$.

\section{Methodology and Theoretical Framework}

The discussion in the previous section has focused on representing the geometrical configuration of the proposed hybrid antenna array. 
In this section, we aim to describe the theoretical methodology and backgrounds for enhancing the performance of the proposed hybrid antenna array with bowtie elements for the beamforming application.  
The proposed LS-SVM for the DoA estimation measures outputs of $N$ array elements and predicts the directions of $L$ signals impinging on the array elements, as demonstrated in Fig.~\ref{fig:configuration}. 
The LS-SVM technique employs both the training dataset and received signals to generate the DoA estimation and update weights of array elements. 
The LS-SVM algorithm tries to maximize weights in the desired directions and minimize remaining weights in directions of interference signals in every iteration of weight updating. 

\subsection{Three-class classification LS-SVM Method}

\begin{figure}[ht]
  \centering
  \includegraphics[width=0.75\textwidth]{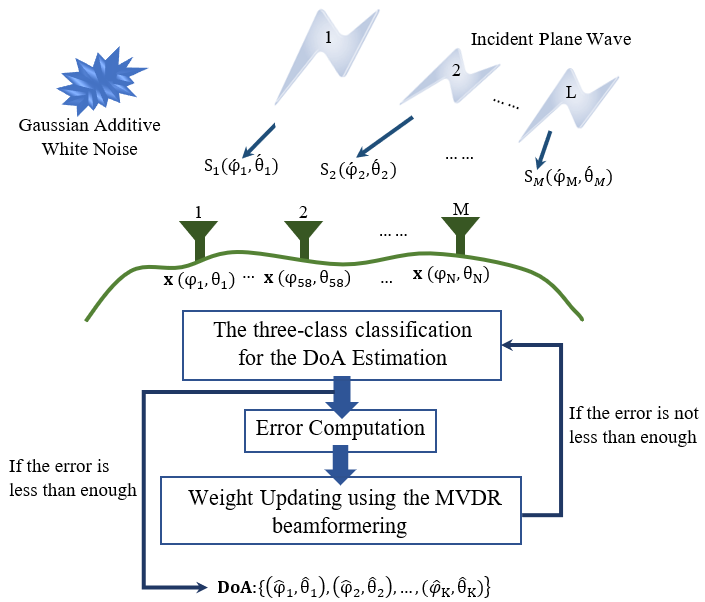}
  \caption{Procedures for the DoA estimation. It is assumed that there are $M$ hybrid antenna array. Each hybrid antenna array consists of 57 elements whose outputs have been measured.}
\label{fig:estimation}
\end{figure}

The LS-SVM algorithm has been employed for changing excitation coefficients for each array element in order to adaptively perform the beamforming and nullsteering technique for the radiation pattern of the hybrid antenna array, as demonstrated in Fig.\ref{fig:estimation}. 
In this section, the method of three-class classification LS-SVM has been described for observing input data and deploying classification rules to generate multi-class labels, and thereby predicting the directions of impinging signals on the hybrid antenna array in the four following procedures. 

\textbf{Methods of Modeling and Producing Data:}
Data modeling is the process of producing data based on assumptions of the real environment. 
In this study, we have employed the MATLAB programming platform to provide raw data. 
Since Mathworks functions of the MATLAB programming platform have been extensively tested, evaluated and verified based on IEEE standards and criterion, thereby simulation results obtained through MATLAB provide a very high level of realistic accuracy.
In the preliminary stage of producing data, a sinusoidal function is considered as an available signal. 
In this scenario, the available signal is affected by the environment noise, the coupling effect, and communication channel conditions. 
In this study, conditions of the communication channel are supposed to be the Ricean fading channel, LoS, and equal correlation matrices, as described in detail in~\cite{Tataria}. 
However, we have explored effects of NLOS and unequal correlation matrix on elements of the hybrid antenna array in section~\ref{sec:performance}.
In addition, the additive white Gaussian noise (AWGN) is added to the simulated signal by the standard MATLAB function AWGN. 
Afterward, the standard MATLAB function \textit{collectPlaneWave()} exerts the plane wave condition to the simulated signal or impinging signals to the hybrid antenna array, which has experienced the AWGN noise and channel conditions.
\textbf{The training procedure:} 
The machine learning algorithm analyzes the training dataset and constructs classifier hyperplanes from the training dataset such that there is a maximum margin between the different classes of data. Therefore, the output is a model trained by the training dataset. 

\begin{equation}
    \mathcal{I} = \{x_i,y_i^c \}_{i=1,p=1}^{i=n,p=P}
    \label{eq:I}
\end{equation}
The training labels are defined in the following equation,
\begin{equation}
    y_i\in\{1,2,\ldots,G\}
\end{equation}									
\noindent where $G$ refers to the output labels. Moreover, Eq.~\ref{eq:I} holds for the $n$ index of training pattern and the $p$ number of classes. Scalar dot products between pairs of inputs and outputs are performed in a higher dimension of the feature space by kernel functions. 
In the feature space of a higher dimension, pairs of inputs and outputs can be separated by linear hyperplanes. 
$p$ classes are mapped by the standard basis into the $\mathbb{R}^p$ space. 
Consider an input of $x_{i}$ is associated with $p$th class, $x_{i} \in C_{p}$. Then, its corresponding output of $y_{i}$ is mapped by a binary row vector of $1$ in the $p$th position and $0$ in all other positions,
\begin{equation}
    x_{i} \in C_{p} \rightarrow y_{\text{ij}}=\left\{ \begin{matrix}
1\ \ \ j = p \\
0\ \ \ O.W. \\
\end{matrix} \right.
\end{equation}

\noindent Hyperplanes of the multi-class classifier, which are normal to their corresponding weights, can be expressed by,
\begin{equation}
\text{ \ \ \ \ }{\text{\ \ \ }x}_{\ } \in C_{p} \rightarrow
\\p=\argmax g_{j}(x)
\label{eq:argmax}
\end{equation}
\noindent for $j=1,2, \ldots, p$
where $g_{J}\textbf{(x)}$ refers to the non-linear softmax function in Eq.~\ref{eq:argmax}.
\begin{equation}
    g_{j}(x) = \frac{exp( < \varphi\textbf{(x)},\textbf{w}^{(j)} > + \textbf{b}_{j})}{\sum_{i = 1}^{k}{exp( < \varphi\textbf{(x)},\textbf{w}^{(j)} > + \textbf{b}_{j})}}
\end{equation}

\noindent where $\textbf{w}^j$ indicates a vector of weights corresponding to a class of $j$. 
Therefore, for an input data of 
\begin{equation}
\{(x_i,y_i)|x_i \in \mathbb{R}^M, y_i \in \mathbb{R}^p,i=1,2,\ldots,N\}, 
\end{equation}
the LS-SVM algorithm is expressed by the following expressions~\cite{Rohwer}, 

\begin{equation}
    \min_{\textbf{w},\textbf{b},\epsilon}{S(\textbf{w},\textbf{b},\epsilon) = \frac{1}{2}}\sum_{j = 1}^{k}{\left(\left\| \textbf{w}^{(j)} \right\|^{2} + \textbf{w}\sum_{i = 1}^{N}\epsilon_{\text{ij}}^{2}\right)}
    \label{eq:ls-svm}
\end{equation}

\noindent where  $\textbf{w}_{ij}$ and $b_j$ refer to an approximate error and a bias term, respectively. In other words. The objective function of $S$ is expressed in terms of a regularizer parameter of $\gamma$ and a superposition of the least squared errors.
An alternative way to solve the proposed optimization method is to obtain its Lagrangian function in Eq.~\ref{eq:lagrangian} and minimize it. 
\begin{equation}
    \mathcal{L}(\textbf{w},\textbf{b},\epsilon,a) = S(\textbf{w},\textbf{b},\epsilon) -\sum_{i = 1}^{N}\sum_{j = 1}^{k}{a_{\text{ij}}\left\lbrack < \varphi\left( \textbf{x}_{i} \right),\textbf{w}^{(j)} > + \textbf{b}_{j} + \epsilon_{\text{ij}} - y_{\text{ij}} \right\rbrack}
    \label{eq:lagrangian}
\end{equation}

\noindent where $a_{ij}$ refers to the Lagrange multipliers in the space of $\mathbb{R}$.
By employing inequality constraints, minimum values of the Lagrangian function can be obtained in the given equation, 

\begin{equation}
    \min_{\textbf{w},\textbf{b}}{\max_{\alpha_{\text{ij}} \geq 0}{\mathcal{L}(\textbf{w},\textbf{b},\epsilon,a)}} =
    \min_{\textbf{w},\textbf{b}}{\max_{\alpha_{\text{ij}} \geq 0}\frac{\left\| \textbf{w}^{(j)} \right\|^{2}}{2}} - \sum_{i = 1}^{N}{a_{\text{ij}}\ \left\lbrack y_{\text{ij}}\left( \textbf{w}^{(j)}\textbf{x}_{i} + \textbf{b}_{j} \right) - 1 \right\rbrack}
\end{equation}

In addition to constructing decision boundaries or classifier hyperplanes during the training procedure, hyperparameters of $b$ and $\gamma$ are updated so that misclassification errors of $\epsilon_{\text{ij}}$ are minimized. 
In other words, classifier hyperplanes are changed by the justification in hyperparameters to minimize mis-classification errors.

Furthermore, the MVDR beamforming technique is implemented to optimize the array pattern by adjusting the weights of array elements so that SINR values are maximized, as described in the next section.  

\textbf{The testing procedure:}
In the testing stage, the remainder of the data set, which was not applied for the training dataset and is called the testing dataset, is fed to the model. As soon as the testing dataset is fed to the model, the model becomes fixed such that it cannot change. 
The model predicts which class each data point of the testing dataset belongs.

\textbf{Model Validation:} 
After the training procedure, to assess the effectiveness and strength of the obtained model to process new data and perform accurate predictions, we have employed the well-established technique of $\mathscr{K}$-fold cross validation~\cite{Kamble2022MachineLA,han2012mining,Belyadi}. 
In this work, it is assumed that $\mathscr{K}$ is identical to five.  
The dataset is portioned into the five subsets or folds with an equal size. In each fold, we have chosen one set as a validation set and the remaining four sets as a training dataset. The reduced training dataset, or four sets, is trained by various hyperplanes. We have obtained the validation accuracy for each fold and the superior model is associated with a fold with the highest accuracy. 
Then, the superior model is trained by the whole of the training dataset including the reduced training dataset and validation set. 
This results in the ultimate model. The generalization error is obtained by evaluating this ultimate model with the testing dataset.  
In this work, the size of the training dataset and the testing dataset is equal to $56\times67$, and the given dataset has randomly been split into 75\% for the training dataset and 25\% for the testing dataset. 
Furthermore, the regularization parameter of $\gamma$ is equal to 10.

\subsection{The DoA Estimation}

The discussion in the previous section has focused on describing the multi-class classification LS-SVM method. In this section, the LS-SVM method has been implemented for the DoA estimation. 
The LS-SVM method has to be capable of generating multi-class labels of $y_i\in\{1,2,\ldots,G\}$ for $L$ dominant signal paths, where $\mathbf{x} \in [-90^\circ \\ 90^\circ]$
as demonstrated in Fig.~\ref{fig:estimation}. 
The multi-class labels results from a set of 
$x \in [S_i]$,
in which $S_{i}$ or the objective function of Eq.~\ref{eq:ls-svm} represents the field of view for $i$th array element.     

Beamforming is employed for calculating the scalar product between the measured data at outputs of array elements and the steering vector in the following equation, 
\begin{equation}
    | \braket{\mathscr{R}_n(\phi_{l},\theta_{l}),\mathbf{h}_n(\phi_{l},\theta_{l})}|^2
\end{equation}
\noindent 
\hspace{8mm}for $l=1, 2,\ldots,L$\\
\hspace{8mm}and $n=1, 2,\ldots,N$.\\
Eq.14 represents the mean of the estimated power of the source.

\begin{equation} 
    \mathscr{R}_n(\phi,\theta) = \sum_{l = 1}^{L} \left(
     \underbrace{\mathbf{C}_n e^{j \mathbf{K}.\mathbf{r}_n}}_{\text{source}} + \underbrace{\mathbf{v}_n}_{\text{noise}}
    \right)
\end{equation}

\noindent where $\mathbf{v}_n = \sigma^2 \mathbf{I}$ for $n = 1,2,\ldots,N$.

The steering vector of the proposed hybrid antenna array in Fig.~\ref{fig:configuration} is obtained by the multiplication of the steering vector matrices of the three cylindrical antenna arrays and circular antenna array in the following equations~\cite{Tan}, 

\begin{equation}
    h_n^{(T)}(\phi,\theta) = \underbrace{\frac{1}{N^{(1)}} \frac{\mathbf{C}_n^{(1)}(\phi,\theta)}{|\mathbf{C}_n^{(1)}(\phi,\theta)|} \otimes
    \frac{1}{N^{(2)}} \frac{\mathbf{C}_n^{(2)}(\phi,\theta)}{|\mathbf{C}_n^{(2)}({\phi,\theta})|}\otimes
    \frac{1}{N^{(3)}} \frac{\mathbf{C}_n^{(3)}({\phi,\theta})}{|\mathbf{C}_n^{(3)}({\phi,\theta})|}}_{\text{for the three cylindrical antenna arrays}}\otimes
    \underbrace{\frac{1}{N^{(4)}} \frac{\mathbf{C}_n^{(4)}({\phi,\theta})}{|\mathbf{C}_n^{(4)}({\phi,\theta})|}}_{\text{for the circular antenna array (when h=0)}}
\end{equation}

\noindent in which the two-dimensional steering vector of each cylindrical antenna array in terms of phase differences or coefficients of the source vector is given by,

\begin{equation}
    \textbf{h}_n (\phi,\theta) = \frac{1}{N} \frac{\mathbf{C}_n(\phi,\theta)}{|\mathbf{C}_n(\phi,\theta)|} 
\end{equation}

\noindent where coefficients of the source vector are expressed by,
\begin{equation}
\mathbf{C}(\phi,\theta) = 
\left[
\begin{matrix}
    g_1(\phi,\theta) e^{-\frac{j2\pi}{\lambda}(r \sin \theta \cos(\theta - \theta_1) + h \cos \phi)} \\
    g_2(\phi,\theta) e^{-\frac{j2\pi}{\lambda}(r \sin \theta \cos(\theta - \theta_2) + h \cos \phi)}\\
    \vdots \\
    g_N(\phi,\theta) e^{-\frac{j2\pi}{\lambda}(r \sin \theta \cos(\theta - \theta_N) + h \cos \phi)}
\end{matrix}
\right]
\end{equation}

\begin{equation}
    \mathbf{K} = \frac{2\pi}{\lambda}(K_x,K_y,K_z) = (\sin \phi \sin \theta, \sin \phi \cos \theta, \cos \phi)
\end{equation}

\noindent where $\mathbf{K}$ represents the wavelength vectors in directions of radius vectors of $\mathbf{r}_n$ pointed to $n$th array elements.

\begin{equation}
    \mathbf{r}_n^T = [r \cos \theta_n, r \sin \theta_n, h]^T
\end{equation}
\noindent Hence, the phase shift relative to the origin is given by, 
\begin{equation}
    \mathbf{K}.\mathbf{r}_n = \frac{2\pi}{\lambda} [\sin \phi \sin \theta \hspace{4mm} \sin \phi \cos \theta \hspace{4mm} \cos \phi]
    \left[
    \begin{matrix}
    r \cos \theta_n \\ r \sin \theta_n \\ h
    \end{matrix}
    \right] = \frac{2\pi}{\lambda}(r \sin \phi \cos \theta \cos \theta_n + r \sin \phi \cos \theta \sin \theta_n + h \cos \phi)
\end{equation}
Furthermore, $g_n(\phi,\theta)$ refers to the antenna gain at the $n$th element.
\begin{figure}[htbp]
\centering
(a)
\includegraphics[width=0.58\textwidth,height=0.3\textwidth]{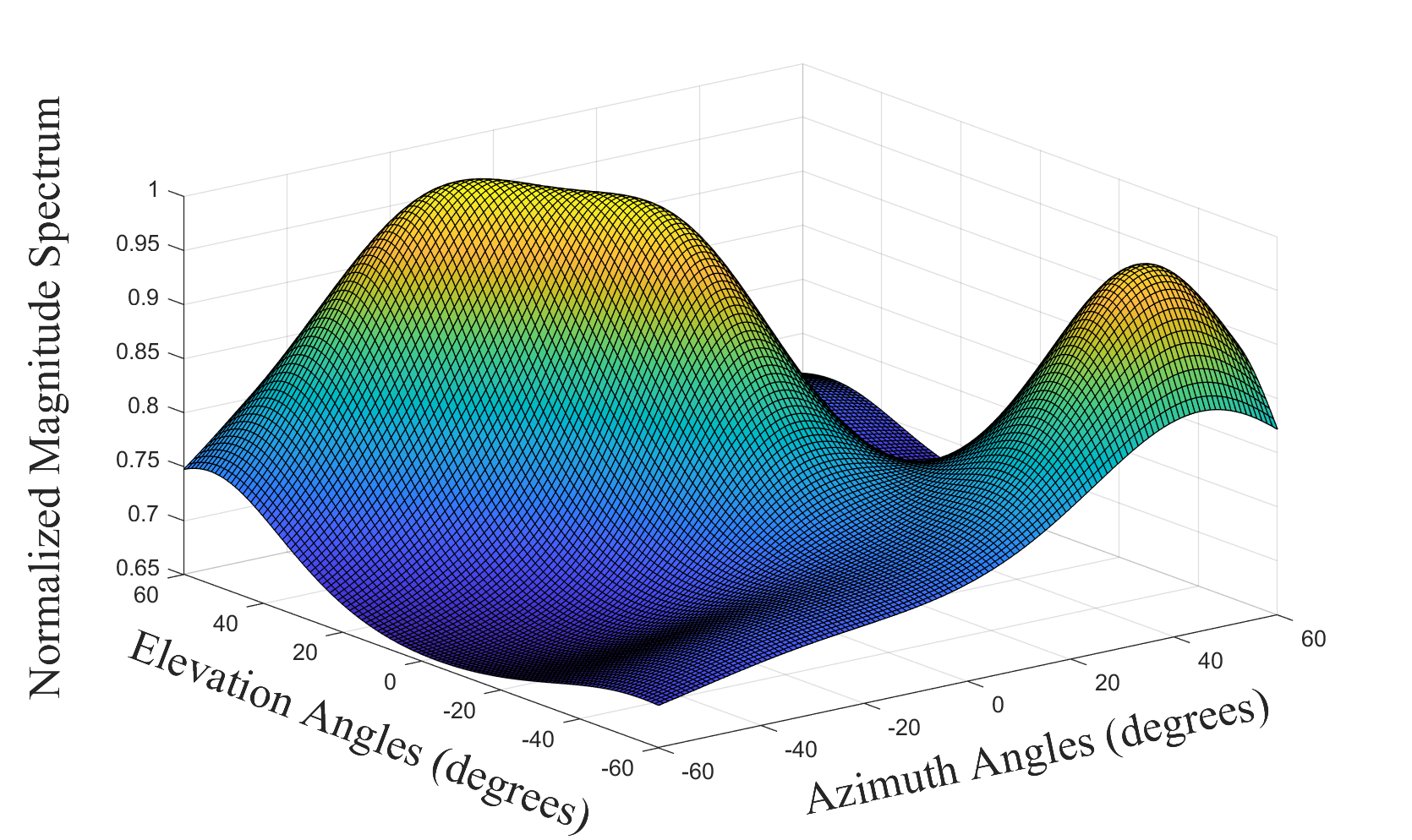}
\\(b)
\includegraphics[width=0.58\textwidth,height=0.3\textwidth]{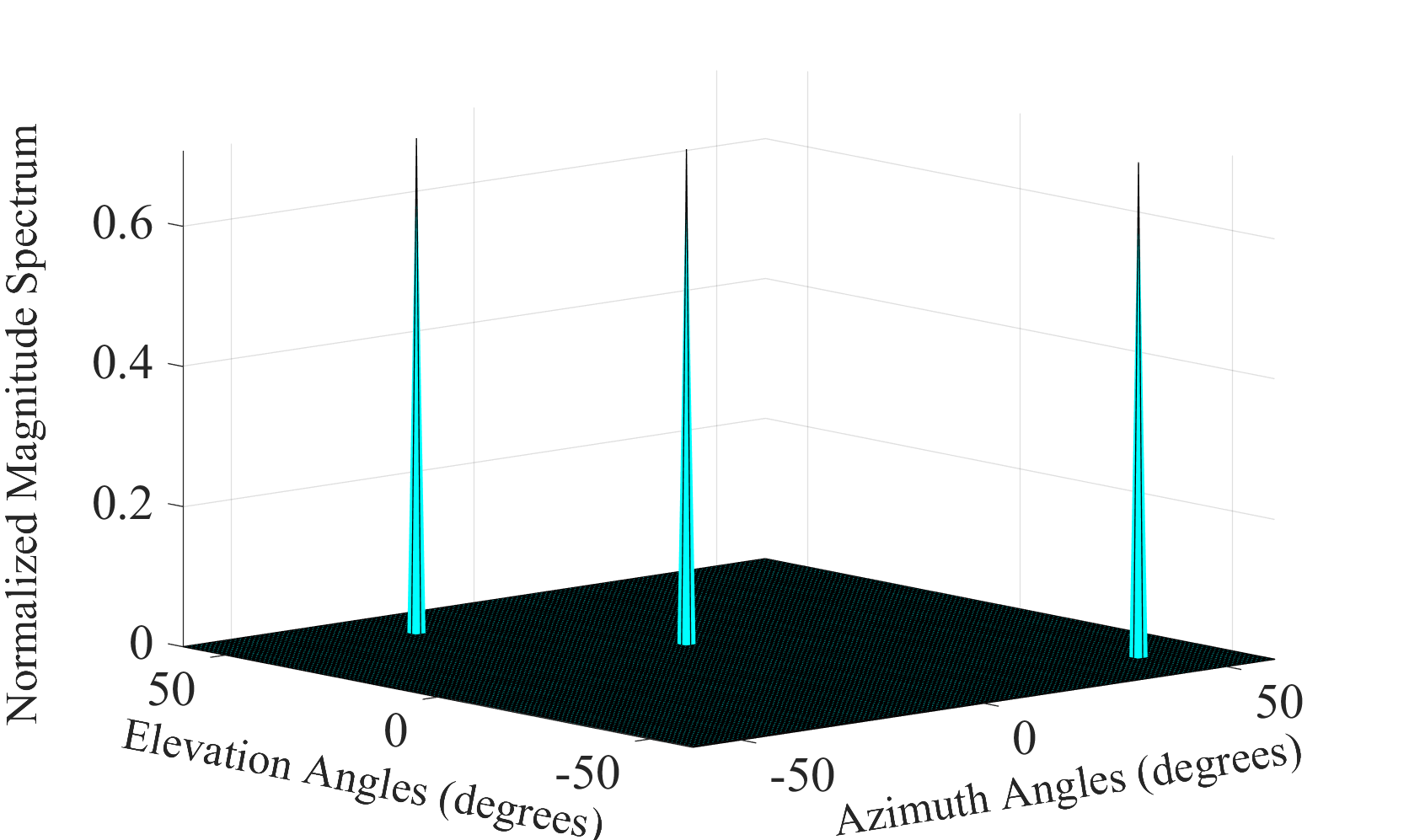}
\caption{The 3D DoA estimation at the frequency of 10 GHz. (a) without performing any optimization method, and (b) with performing the three-class classification LS-SVM model for the hybrid antenna array with bowtie elements, consistent with Fig.1(b).}
\label{fig:doa-estimation_svm}
\end{figure}
Fig.~\ref{fig:doa-estimation_svm} has reported the numerical results of implementing the three-class classification LS-SVM method for the hybrid antenna array with bowtie elements for the DoA estimation. 
Although, the DoA estimation allows for determining the dominant signal sources, the main practical challenge resides in estimating the low-amplitude signal sources. 
The implementation of the three-class classification LS-SVM method for the hybrid antenna array with bowtie element is capable of significantly enhancing the performance of the DoA estimation, as demonstrated in Fig.~\ref{fig:doa-estimation_svm}.   

\subsection{MVDR Beamforming technique for Updating Weights}

Parameters of a wireless communication channel have to vary quickly, thereby weight updating should be performed at a higher rate than alternative. 
An alternative solution for updating the weight vector consists of directly computing the inverse of the steering vector using the MVDR algorithm. 
The primary goal of the MVDR beamforming technique is to minimize the signal to SINR expressed by the following formula~\cite{Darzi,Vorobyov},

\begin{equation}
    \text{SINR} \triangleq \frac{{| \braket{(\mathbf{C}_k e^{j \mathbf{K}.\mathbf{r}_k}),(\mathbf{h}(\phi_{k},\theta_{k})}|^2}}
    {{| \braket{(\mathbf{C}_k e^{j \mathbf{K}.\mathbf{r}_k} + \mathbf{v}_k +\mathbf{i}_k),(\mathbf{h}(\phi_{k},\theta_{k})}|^2}}
    \label{eq:sinr}
\end{equation}
\noindent Hence, the minimum SINR is achieved by the MVDR beamforming technique when the following conditions are satisfied,

\begin{align}
    & \min_\alpha \braket{(\mathbf{C}_k e^{j \mathbf{K}.\mathbf{r}_k} + \mathbf{v}_k +\mathbf{i}_k),(\mathbf{h}(\phi_{k},\theta_{k})}\\
    & \text{ s.t.} \\
    & \hspace{7mm} (\mathbf{h}^H(\phi_{k},\theta_{k},
    \textbf{C}_k(\phi,\theta)=1
\end{align}
\noindent Hence, the MVDR beamforming solution is expressed,
\begin{equation}
    \textbf{w}_{(MVDR)} =\alpha\mathbf{h}^{-1}({\phi_{-n+k},\theta_{-n+k})}|^2 \textbf{C}_k(\phi,\theta)
    \label{eq:MVDR}
\end{equation}

\noindent The normalized constant of $\alpha$ in Eq.~\ref{eq:MVDR}, which does not have any effect on the SINR, can be omitted.
\begin{equation}
\alpha = \frac{1}{{| \braket{(\mathbf{C}_k),(\mathbf{h}^{-1}({\phi_{-n+k},\theta_{-n+k}}))}|^2}} 
\end{equation}
However, in practical scenarios, weight updating performance is limited by impossibility of measuring $R_{n+k}$. 
Hence, we can employ a sample covariance matrix of $R$, or an auto-correlation matrix of $R$, which is expended in terms of a summation of the multiplication of the vector of the input data signals, $r_A (k)$, and its transpose matrix in the following equation,
\begin{equation}
    \hat{\textbf{R}} = \frac{1}{M} \sum_{k=K-M+1}^K r_A (k)  r_A^H (k)
\end{equation}

\section{Performance Evaluation}
\label{sec:performance}

The discussion in the previous sections has concentrated on verifying the beamforming performance of the hybrid antenna array for the DoA estimation. 
In this section, we aim to evaluate the beamforming performance of the hybrid antenna array in terms of SINR, antenna efficiency, and the 3D uniform beam-scanning capability in wireless communication channels. 

\subsection{Signal to Interference and Noise Ratio}
SINR values, Eq.~\ref{eq:sinr}, represent the capability of the antenna array in suppressing interference and overcoming the noise of wireless communication channels. 
The capability of the adaptive antenna array to minimize the error between the desired and actual signals, and thereby to maximize SINR values, is controlled by the two underlying factors of geometrical parameters, such as the weight (steering) vector of $h{(\textbf{x})}$ and inter-element spacing, and conditions of the wireless communication channels.
\textbf{The Effect of Geometrical Parameters of the Antenna Array:}
The spatial correlation between array elements at a cellular base station (BS) is very much affected by the average angles of directions of arrival signals, inter-element spacing, and angle spread. 
In addition to the aforementioned factors, the mutual coupling effect causes changes in eigenvalues of the signal covariance matrix~\cite{Yuan} and hence affects the transient response of the hybrid antenna array. 
Particularly, when the inter-element spacing decreases, the mutual coupling effect becomes stronger, which is accompanied by lower eigenvalues, and thereby a longer transient response. 
Therefore, the speed of the antenna array response includes delays in suppressing jammers~\cite{Mandric,Singh} and interference signals. 
Hence, the output of SINR represents more transient oscillatory responses in the presence of the mutual coupling effect compared with the absence of mutual coupling effect and thereby a significant degradation in SINR values. 
In this scenario, Xiaojian Wang in~\cite{Wang} has verified that the available solutions to implement and develop the compensations for mitigating the loss of SINRs as well as for increasing the speed of the antenna array response are not efficiently capable of overcoming the aforementioned constraints. 
He has also demonstrated that the optimization techniques have significantly enhanced the efficiency of the array synthesis process. Following this concept, in this study, we have drastically improved the SINR performance by deploying the proposed LS-SVM optimization method for the hybrid antenna array with bowtie elements.  

\textbf{The Effect of Conditions of the Wireless Communication Channels}

Another significant challenge of deploying antenna arrays for beamforming techniques for the behind 5G and 6G technologies of wireless communication channels consists of the type of channel fading. Although several different techniques have been developed for minimizing SINRs in massive wireless communication channels~\cite{Tataria,su12219006}, limited literature has studied the effects of LoS and NLoS and inequality in the spatial correlation, especially, when the size of the antenna arrays reduces significantly. 
Furthermore, in practice, when a different set of directions of the impinging signal is observed by multiple terminals of array elements or antennas, unequal correlation matrices have been observed by array terminals.
Hence, in this section, the effects of the wireless communication channels on the performance of the hybrid antenna array have been studied when the proposed LS-SVM algorithm has been performed.

Consider an uplink of the wireless communication channel, in which a circular cell is allocated for each BS. 
Each BS is equipped with $M$ numbers of the hybrid antenna arrays which communicate with $N$ numbers of terminals of array elements $(N\gg M),$ as demonstrated in Fig.~\ref{fig:estimation}. 
The received signal including an $M\times 1$ matrix at each BS is expressed by,
\begin{equation}
\mathscr{R}= \rho^{1/2}\textbf{G} \textbf{D}^{1/2}\textbf{S} +\mathbf{v}_n
\end{equation}
    
\noindent where $\rho$ refers to the average uplink transmitted power. 
$\textbf{G}$ denotes an $M\times N$ matrix of the fading channel between the $M$ numbers of BS antennas and $N$ numbers of terminals of array elements. 
$\textbf{D}$ indicates an $N\times N$ diagonal matrix of link gains. The link gain of $n^{ \text{th}}$ terminal is given by,
\begin{equation}
    [\textbf{D}]_{n,n}=\beta_n.
\end{equation}

\noindent Thus, SINR can be expanded for $n^{ \text{th}}$ terminal by the following formulation,
\begin{equation}
    \text{SINR}_n = (\rho \beta_n  \|\textbf{y}_n \|^4)/(\|\textbf{y}_n \|^2+\rho \sum_{k\neq n, k=1}^N \beta_k  |\textbf{y}_n^H  \textbf{y}_k |^2) 
\end{equation}

\noindent 
\begin{equation}
    \textbf{y}_n=\eta_n \textbf{q}_n + \gamma_l \textbf{R}_n^{1/2} \textbf{q}_n
\end{equation}

\noindent $\textbf{q}_n$ and $\textbf{q}_n$ refer to the LoS and NLoS components of the channel fading. 
$\eta_n$ and $\gamma_n$ denote coefficients of $k^{ \text{th}}$ factor of the channel fading at the $n^{ \text{th}}$ terminal.
$\mathscr{R} R_l$ represents the received correlation matrix for $n^{\text{th}}$ terminal. We did not expand the theoretical and mathematical computations for estimating SINRs in details~\cite{Tataria,su12219006}. 

\begin{table}[ht]
    \centering
    \begin{tabular}{P{2.0cm}|P{3.0cm}|P{2.0cm}|P{3.0cm}}
       Element type  &  LoS, equal correlation & NLoS, equal correlation & NLoS, unequal correlation \\ \hline
       Bowtie elements  & 35.218 (dB) & 33.858 (dB) & 31.564 (dB) \\
       Dipole elements & 20.630 (dB) & 19.256 (dB) & 17.368 (dB) \\
    \end{tabular}
    \caption{SINR values when the three-class classification LS-SVM method has been implemented for the hybrid antenna array with bowtie elements and dipole elements, consistent with Fig.1.}    \label{tab:sinr}
\end{table}
\textbf{Channel Condition 1:}
SINR has been obtained assuming the availability of the Rician channel model~\cite{Tataria} with small-scale fading~\cite{Kaur} and the Line-of-Sight (LoS) propagation~\cite{ZhangLAWP} with the equal correlation matrices~\cite{ZhangLAWP}. 
The equal correlation matrices in the presence of the LoS propagation model result from a set of incident waves in the same direction, as discussed in detail in~\cite{ZhangLAWP}. 
In other words, it is assumed that the LoS propagation is performed at array elements of the hybrid antenna array with an equal correlation matrix.  

\textbf{Channel Condition 2:}
The LS-SVM method has been implemented for the hybrid antenna array with bowtie elements in the wireless channel with a pure NLoS propagation (i.e., Rayleigh fading) and unequal correlation matrices at the array elements. The following numerical solutions have been employed for simulating these conditions of the wireless channel. 

\begin{table}[]
    \centering
    \begin{tabular}{c|c} \hline
        Parameter & Numerical values\\ \hline
        $f_s$ & $10^9$ Hz \\
        Path delays & $[0, 200, 800, 1200, 2300, 3700] \times 10^{-9}$ seconds \\
        Average path gains & $[0, -0.9, -4.9, -8, -7.8, -23.9]$ dB\\
        Maximum doppler shift of $f_d$ & 50 Hz
    \end{tabular}
    \caption{Numerical results for the NLOS Rayleigh channel fading with the non-equal correlation matrix.}
    \label{tab:parameters-condition-2}
\end{table}


\textbf{Channel Condition 3:}
The LS-SVM algorithm has been deployed for the hybrid antenna array with bowtie elements in the pure Rayleigh fading channel with equal correlation matrices at the array elements,  
Numerical solutions to simulate the aforementioned conditions of the wireless communication channel have been represented in the following formulation, 

\begin{table}[]
    \centering
    \begin{tabular}{c|c} \hline
        Parameter & Numerical values\\ \hline
        $f_s$ & $10^9$ Hz \\
        Path delays & $[0, 0, 0, 0, 0, 0] \times 10^{-9}$ seconds \\
        Average path gains & $[1, 1, 1, 1, 1, 1]$ dB\\
        Maximum doppler shift of $f_d$ & 50 Hz
    \end{tabular}
    \caption{NLOS Rayleigh channel fading with the equal correlation matrix.}
    \label{tab:parameters-condition-3}
\end{table}


Table~\ref{tab:parameters-condition-3} has represented the analogous results of the SINR performance of the hybrid antenna array in the three different wireless channels of LoS, NLoS with an equal correlation matrix, and NLoS with an unequal correlation matrix. 
A drastic difference of the SINR value of more than 14 dB is noticeable in the hybrid antenna array with bowtie and dipole elements in all three fading channels.

In all three different wireless channels, a drastic enhancement in the SINR values of more than 14 dB have been achieved for the hybrid antenna array with bowtie elements compared to its geometrical counterpart with dipole elements. 
This extreme difference results from the effectiveness and superiority of implementing bowtie elements for the hybrid antenna array over dipole elements.
Furthermore, the results of table~\ref{tab:parameters-condition-3} ensure that the multi-path effect in the NLoS channel degrades the antenna performance with the assumption of an unequal correlation matrix at the array elements significantly compared to the other wireless channels.

\subsection{Radiation Pattern Efficiency}
The high antenna efficiency delivers reliable and much more information, however, the antenna efficiency does not represent any information about radiation directivity. 
In the scenario of the radiation directivity, specifically, the antenna determines to radiate with a greater efficiency towards a particular direction of its surrounding space. 
The antenna efficiency is highly sensitive to technological imperfections, especially the mutual coupling effect, and therefore surface waves.
The antenna efficiency is expressed in terms of the efficient radiated power, $P_{rad}$, the power which leaks at the back of the array, $P_{back}$, and the surface-wave power which traps in the planes of sub-arrays, $P_{sw}$, in the following formula,

\begin{equation}
\eta_e=\frac{P_{rad}}{(P_{rad}+P_{sw}+P_{back})} 		
\end{equation}
\noindent where $P_{sw} + P_{back} = P_{loss}$

\begin{table}[ht]
    \centering
    \begin{tabular}{P{2cm}|P{3.0cm}|P{3.0cm}}
    & Hybrid antenna array with dipole elements & Hybrid antenna array with bowtie elements \\ \hline
    Antenna efficiency (\%) & 67\% & 92\% \\
    \end{tabular}
    \caption{Antenna efficiency when the three-class classification LS-SVM algorithm method has been performed for the hybrid antenna array with bowtie elements and dipole elements, consistent with Fig.1.}
    \label{tab:efficiency}
\end{table}
Analogously, table~\ref{tab:efficiency} has reported that there is a significant difference in the antenna efficiency of the hybrid antenna array with bowtie elements compared to its geometrical counterpart with the dipole elements, of more than 20\%. 
This practical interest results from the superior performance of the bowtie elements compared to the dipole elements for the hybrid antenna array. 

\subsection{3D Uniform Beam-scanning Capability}
Another significant advantage of the proposed hybrid antenna array with bowtie elements is to allow for uniform 3D-beam scanning.    
The maximum array factor, and thereby the maximum directivity, can be obtained when the entire phase contribution to the array factor is identical to equal unity. 
By satisfying the aforementioned condition in Eq.~\ref{eq:af}, the maximum array factor, in principle, is theoretically obtained by,
$d_v=0.5\lambda$, $d_r=0.5\lambda$, and $(\theta_0,\phi_0 )=(\approx45^{\circ},\approx45^{\circ})$

In this sense, the maximum directivity in terms of the maximum array factor can be derived as follows~\cite{ZhangDOAEF,ijamec277798},
\begin{equation}
D_0=\frac{4\pi |AF(\theta_0,\phi_0)|^2}{(\int_0^2\pi\int_0^\pi|AF(\theta_0,\phi_0)|^2\sin\theta d\theta d\phi}
\end{equation}
\begin{figure}[ht]
(a)
    \centering
    \includegraphics[width=0.58\textwidth]{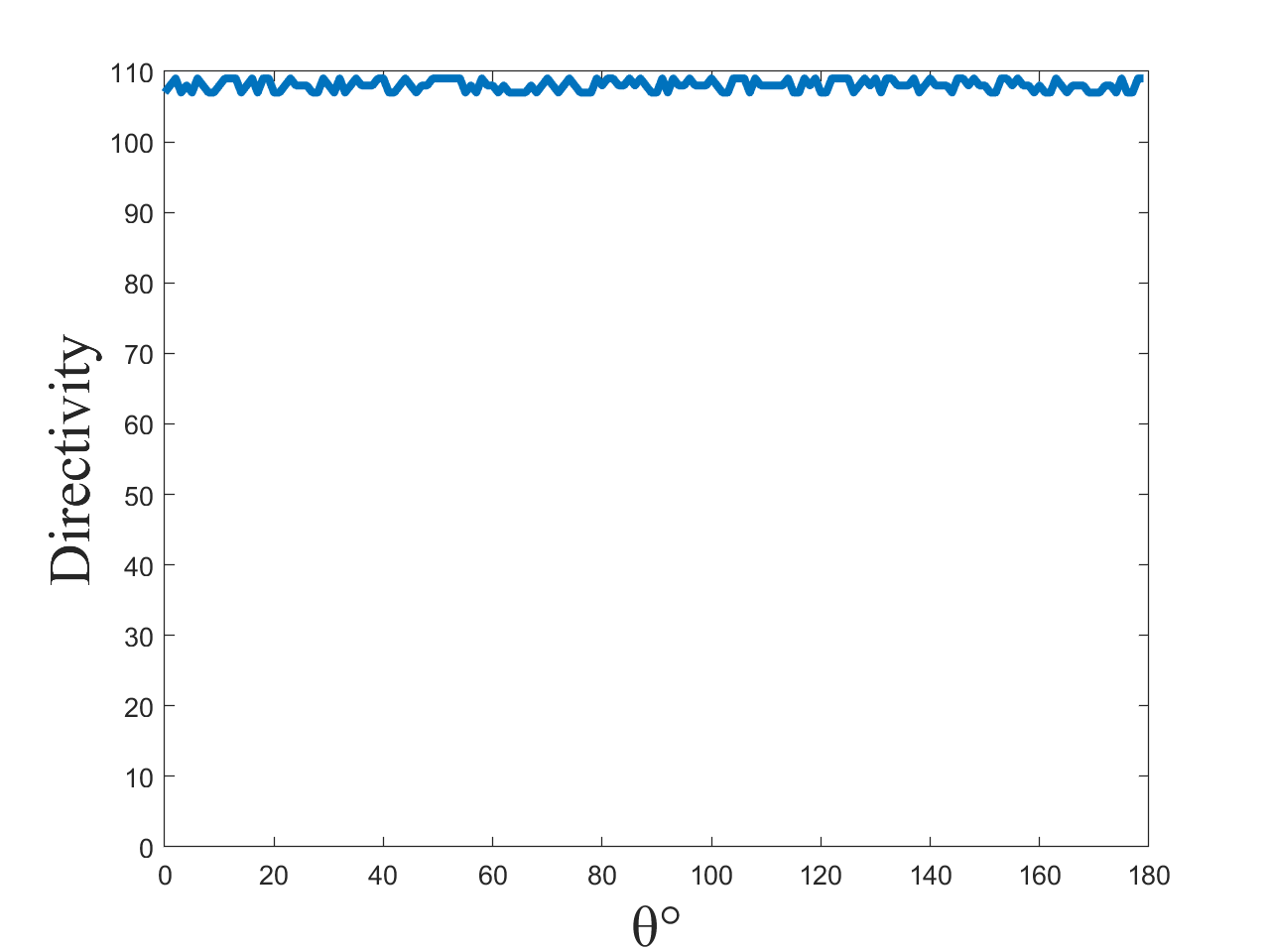}\\(b)
    \includegraphics[width=0.58\textwidth]{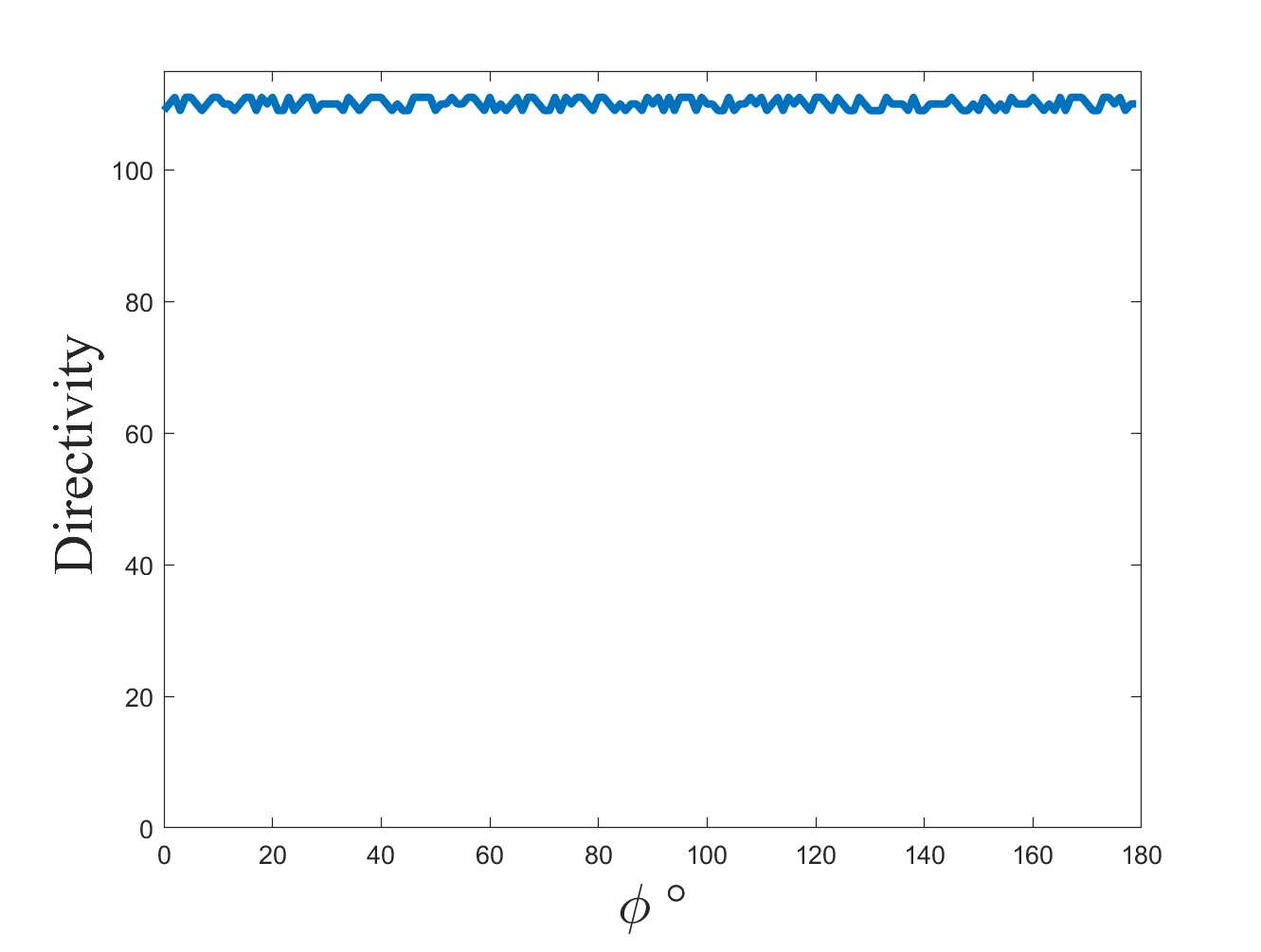}
    \caption{Variation of maximum directivities versus $\theta$ and $\phi$ angles at the frequency of 10 GHz for the hybrid antenna array with bowtie elements consistent with Fig.1(b).}
    \label{fig:directivity}
\end{figure}

The two features can be underlined from Fig.~\ref{fig:directivity}: (1) the hybrid antenna array with the bowtie elements has illustrated that the variation of the maximum directivities remains almost constant throughout the scanning process over $\theta$, while the $\phi$ angle is fixed at $45^{\circ}$ and vice versus.
In addition, the variation of maximum directivities in terms of $\phi$ stays almost invariable at $\theta=45^{\circ}$. Therefore, the hybrid antenna array has the superior performance in terms of the uniform 3D-beam scanning compared to the other different array geometries~\cite{Liu,Balanis-2012-antenna,Yasar,Mahmoud,Ram,Noordin,Harabi,Li,Hawkes,Hussain,Mandric}, (2) the hybrid antenna array with bowtie elements has shown very high maximum directivities~\cite{Ioannides}.

\section{Conclusion}

To conclude, we have outlined some of the current techniques and proposals that implement optimization methods for enhancing the performance of array elements for beamforming applications in wireless communication channels.

The methodology and theoretical framework for the DoA estimation and the beamfoming technique based on the three-class classification LS-SVM method and the MVDR technique, respectively, have been provided. 
We have verified the superior performance of the three-class classification LS-SVM technique in the classification and resolution in Fig.~\ref{fig:doa-estimation_svm}. 

The strong effectiveness and improvement of deploying bowtie elements over dipole elements for the hybrid antenna array have been demonstrated in terms of SINR, antenna efficiency, and the 3D beam-scanning capability. 
This makes the hybrid antenna array with bowtie elements as a potential candidate for the beamforming applications in the wireless communication channels. 

The uniform three-dimensional scanning coverage of the space with the high gain radiation beams has been achieved for the hybrid antenna array with bowtie elements. 
Therefore, the proposed hybrid antenna array is capable of re-orienting its maximum radiation pattern without electronic phase shifting circuitry. 

The hybrid antenna array with bowtie elements has demonstrated SINRs with more than 14 dB compared to its geometrical counterpart with dipole elements. Higher SINR values in the hybrid antenna array with bowtie elements are accompanied by its less sensitivity to variations of wireless communication channels. 

Furthermore, we have verified that the antenna efficiency of the hybrid antenna array with the bowtie elements is about 20\% more than its counterpart configuration with the dipole elements.

\ifCLASSOPTIONcaptionsoff
  \newpage
\fi


\bibliographystyle{IEEEtran}

\bibliography{IEEEabrv,bibtex/bib/antenna}







\end{document}